\documentclass[10pt,twocolumn,letterpaper]{article}

\usepackage{cvpr}      

\usepackage[dvipsnames]{xcolor}

\definecolor{cvprblue}{rgb}{0.21,0.49,0.74}
\usepackage[pagebackref,breaklinks,colorlinks,citecolor=cvprblue]{hyperref}

\def\paperID{*****} 
\def\confName{CVPR}
\def\confYear{2024}

\title{Diffusion360: Seamless 360 Degree Panoramic Image Generation based on Diffusion Models}

\author{Mengyang Feng,~~~Jinlin Liu,~~~Miaomiao Cui,~~~Xuansong Xie\\
Alibaba Group\\
{\tt\small \{mengyang.fmy, ljl191782, miaomiao.cmm, xingtong.xxs\}@alibaba-inc.org}
}

\begin{document}

\twocolumn[{%
\renewcommand\twocolumn[1][]{#1}%
\maketitle
\begin{center}
    \centering
    \captionsetup{type=figure}
    \includegraphics[width=0.95\textwidth]{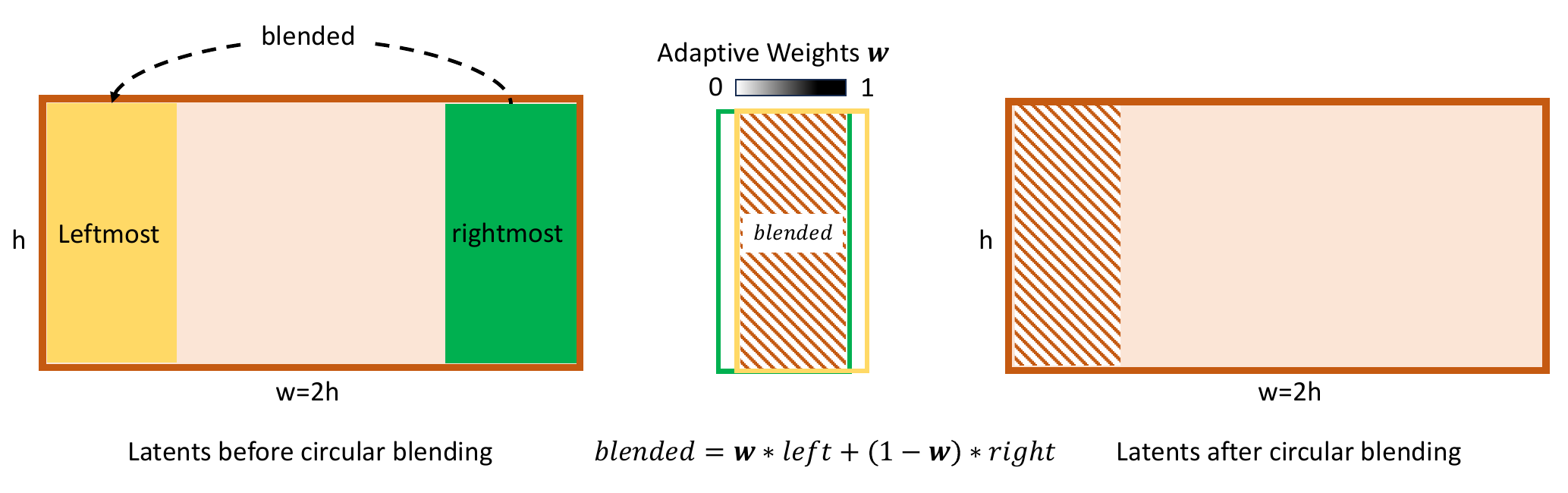}
  \caption{The proposed circular blending operation.}
  \label{circular-blending}
\end{center}%
}]

\maketitle

\begin{abstract}
This is a technical report on the 360-degree panoramic image generation task based on diffusion models. Unlike ordinary 2D images, 360-degree panoramic images capture the entire $360^\circ\times 180^\circ$ field of view. So the rightmost and the leftmost sides of the 360 panoramic image should be continued, which is the main challenge in this field. However, the current diffusion pipeline is not appropriate for generating such a seamless 360-degree panoramic image. To this end, we propose a circular blending strategy on both the denoising and VAE decoding stages to maintain the geometry continuity. Based on this, we present two models for \textbf{Text-to-360-panoramas} and \textbf{Single-Image-to-360-panoramas} tasks. The code has been released as an open-source project at \href{https://github.com/ArcherFMY/SD-T2I-360PanoImage}{https://github.com/ArcherFMY/SD-T2I-360PanoImage} and \href{https://www.modelscope.cn/models/damo/cv_diffusion_text-to-360panorama-image_generation/summary}{ModelScope}.
\end{abstract}

\section{Related Work}
\label{sec:related-work}

Recent studies like MVDiffusion~\cite{Tang2023mvdiffusion}, StitchDiffusion~\cite{wang2023customizing}, and PanoDiff~\cite{zeng2023mirror-nerf} have proved the feasibility of diffusion-based 360-degree panoramic images generation, but still have some drawbacks. 

\textbf{MVDiffusion} needs 8 perspective views (user-provided or generated from Stable Diffusion~\cite{rombach2021highresolution}) as inputs. The resulting closed-loop panoramic image is more like a long-range image with a wide angle. So it has artifacts on the 'sky' and 'floor' when viewing in a 360 image viewer.

\textbf{StitchDiffusion} proposes a global cropping on the left and right side of the image to maintain the continuity. However, it still cracks on the junctions when zoom-in in the 360 image viewer.

\textbf{PanoDiff}, similar to the \textbf{StitchDiffusion}, proposes a circular padding scheme, which is the most related research to our work. The idea of our circular blending strategy is derived from the circular padding scheme. The differences are (1) we use an adaptive weighting policy for geometric continuity, (2) we do not need the \textit{Rotating Schedule} at both training and inference time, which means that we can directly finetune a dreambooth~\cite{ruiz2022dreambooth} model using standard diffusion pipeline for this task, and just apply the circular blending at inference time, and (3) we can directly apply our technique into the ControlNet-Tile~\cite{zhang2023adding} model to produce high-resolution results.


\begin{figure*}
  \centering
  \includegraphics[width=0.6\linewidth]{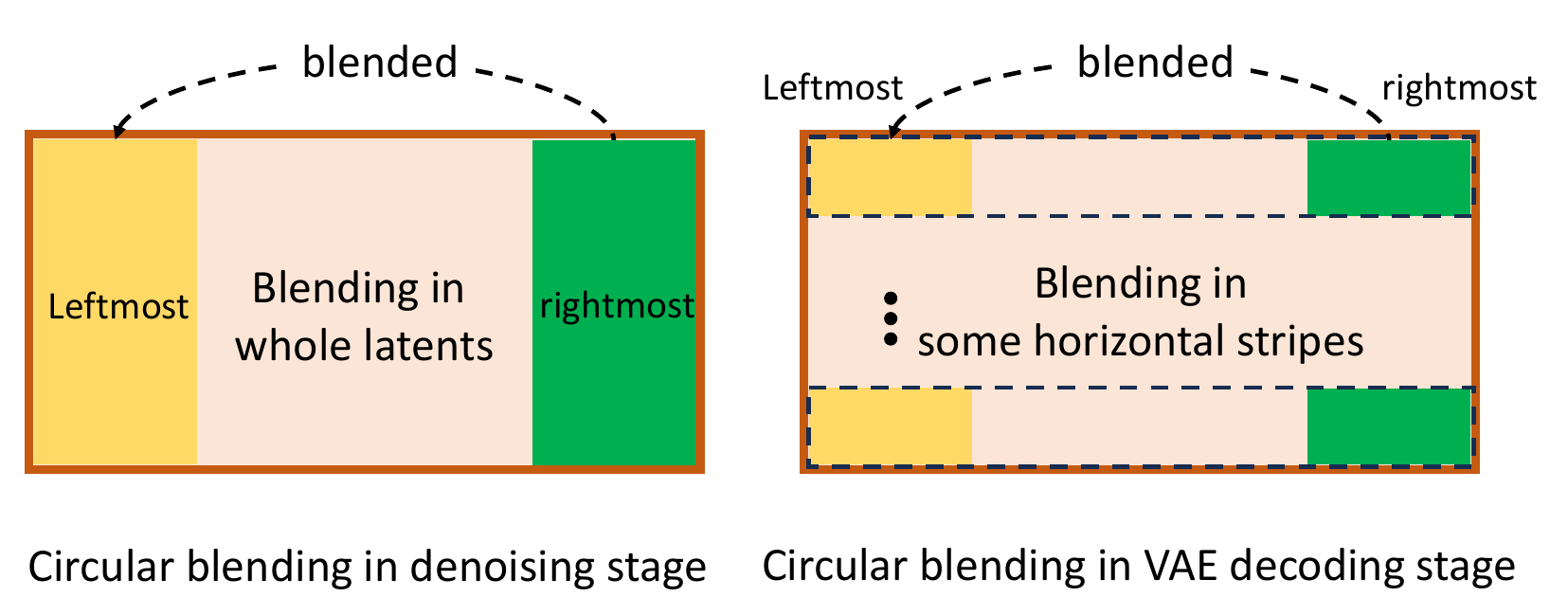}
  \caption{The circular blending operation in different stages.}
  \label{cb-diff-stages}
\end{figure*}

\begin{figure*}[!ht]
  \centering
  \includegraphics[width=0.8\linewidth]{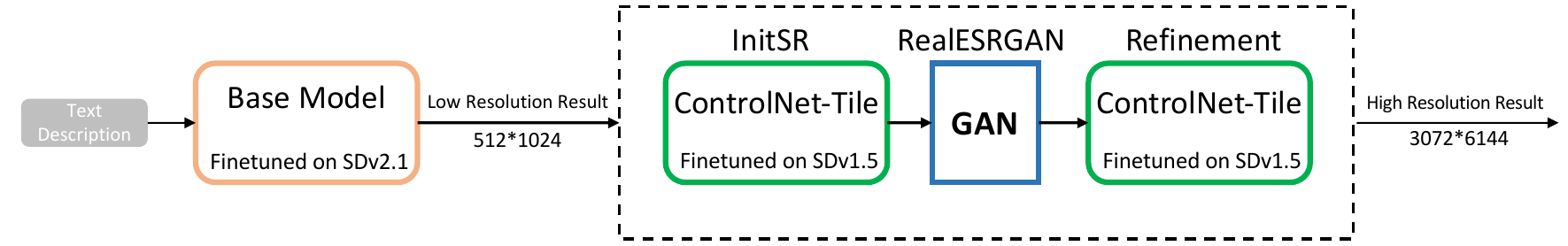}
  \caption{The pipeline of Text-to-360-Panoramas.}
  \label{pipe-text}
\end{figure*}

\section{Method}
\label{sec:method}
\subsection{Circular Blending}
We propose a circular blending strategy at the inference time to generate seamless 360-degree panoramic images. Specifically, at each \textbf{denoising step}, the right part (of a such portion) of the latent feature and the left part (of the same portion as the right part) is blended with adaptive weights. This is illustrated in Fig.~\ref{circular-blending}. Similarly, this strategy can be added to the \colorbox{gray}{\color{white}{tiled\underline{~}decode}} function of the VAE decoder (see Fig.~\ref{cb-diff-stages}). We find that using the circular blending in the VAE decoder is more important than in the latent denoising stage for maintaining the geometric continuity.

\subsection{Text-to-360-Panoramas}

For the Text-to-360-Panoramas task, we propose a multi-stage framework to generate high resolution 360-degree panoramic images. As illustrated in Fig.~\ref{pipe-text}, we first generate a low resolution image using a base model (finetuned on the SUN360~\cite{xiao2012sun360} dataset using the DreamBooth~\cite{ruiz2022dreambooth} training method), and then employ some super-resolution strategies (including diffusion-based and the GAN-based methods, like the ControlNet-Tile model and the RealESRGAN~\cite{wang2021realesrgan}) to up-scale the result to a high resolution one. For better results, we also finetune the ControlNet-Tile model on the SUN360 dataset by generate low-resolution and high-resolution image pairs.

\subsection{Single-Image-to-360-Panoramas}
\begin{figure*}
  \centering
  \includegraphics[width=\linewidth]{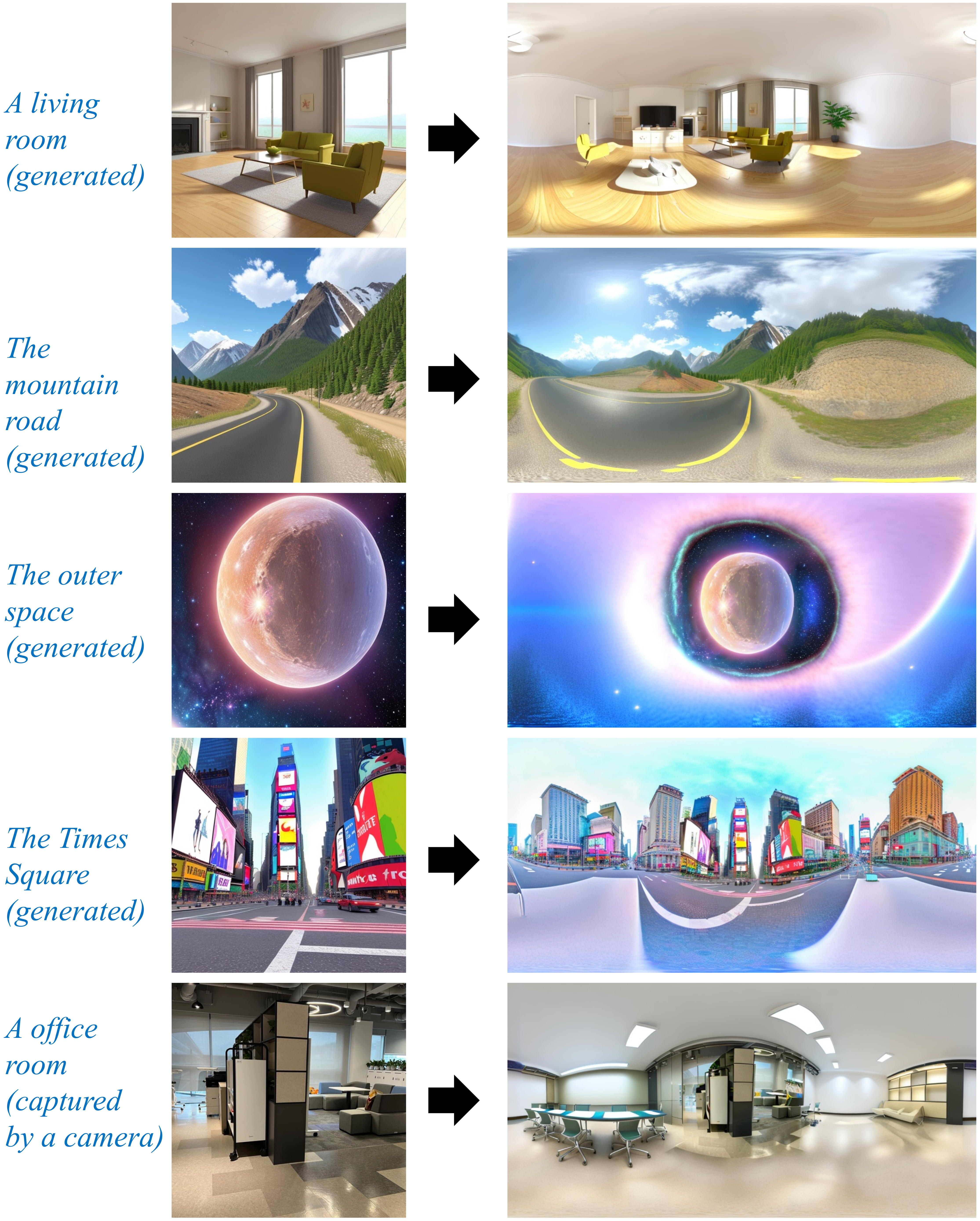}
  \caption{Results of Single-Image-to-360-Panoramas.}
  \label{samples-pto360}
\end{figure*}

For the Single-Image-to-360-Panoramas task, the framework is similar to the Text-to-360-Panoramas by replacing the base model to a controlnet-outpainting model. We design a ControlNet-Outpainting model to generate a low resolution 360-degree panoramic image from a given single ordinary 2D image at perspective view. To generate the training pairs of perspective and panoramic images, we first convert the panoramic image to cube-maps and select the center-cube as its perspective image. The inputs of the ControlNet-Outpainting model consist of the converted center-cube map $C$ with the other cubes filled by zeros and the mask $M$. At inference time, the perspective image can be generated from a certain generative model or captured by a camera (the image should be squared). The perspective image is converted to the center-cube map $C$ as the input of the ControlNet-Outpaining model. For some reason, the trained models of this task can not be released. However, it should be easy to reproduce. See some results in Fig.~\ref{samples-pto360}.

\section{Resuls}
\begin{figure*}
  \centering
  \includegraphics[width=\linewidth]{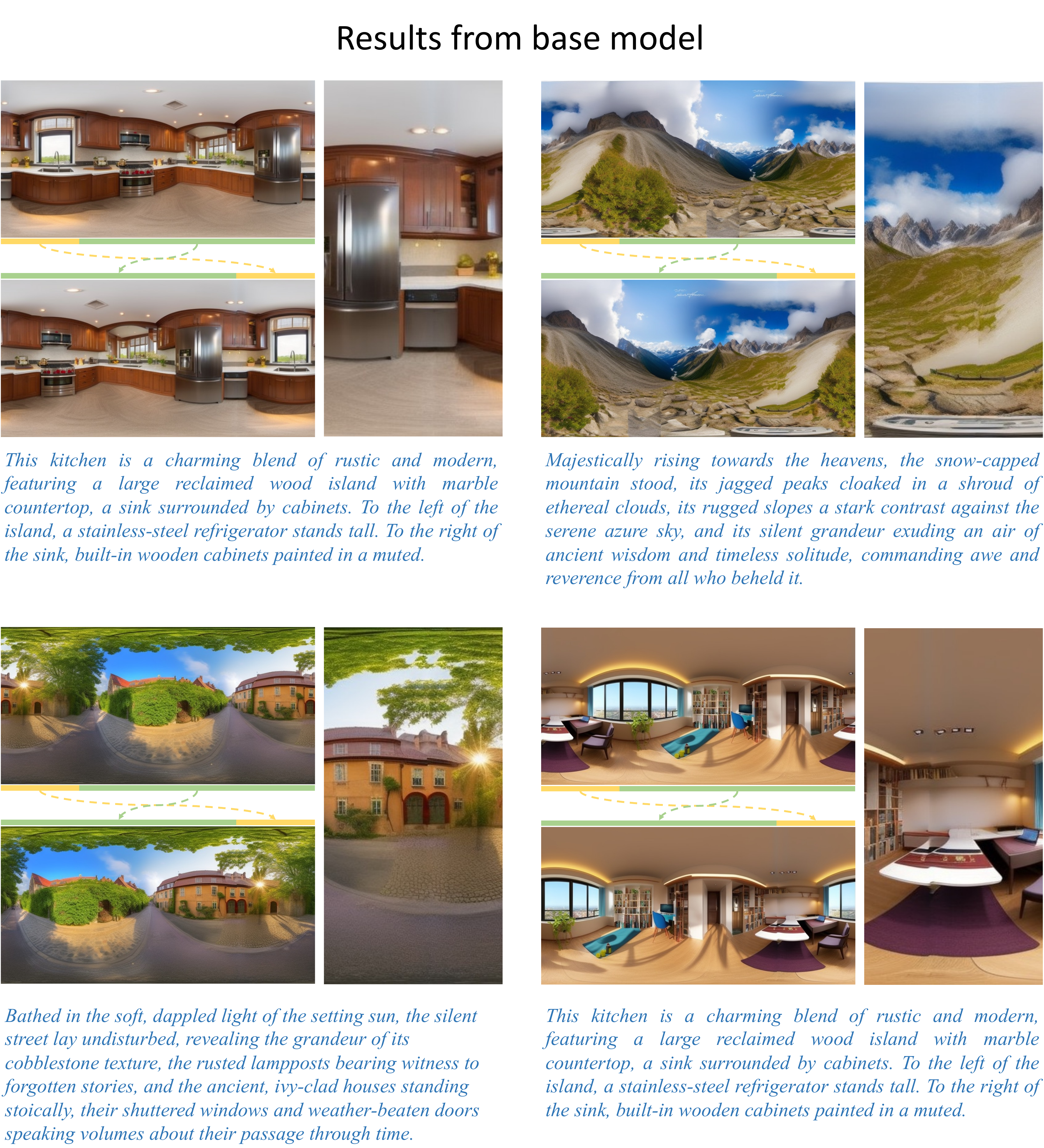}
  \caption{Results from the Base Model.}
  \label{samples-base}
\end{figure*}

\begin{figure*}
  \centering
  \includegraphics[width=\linewidth]{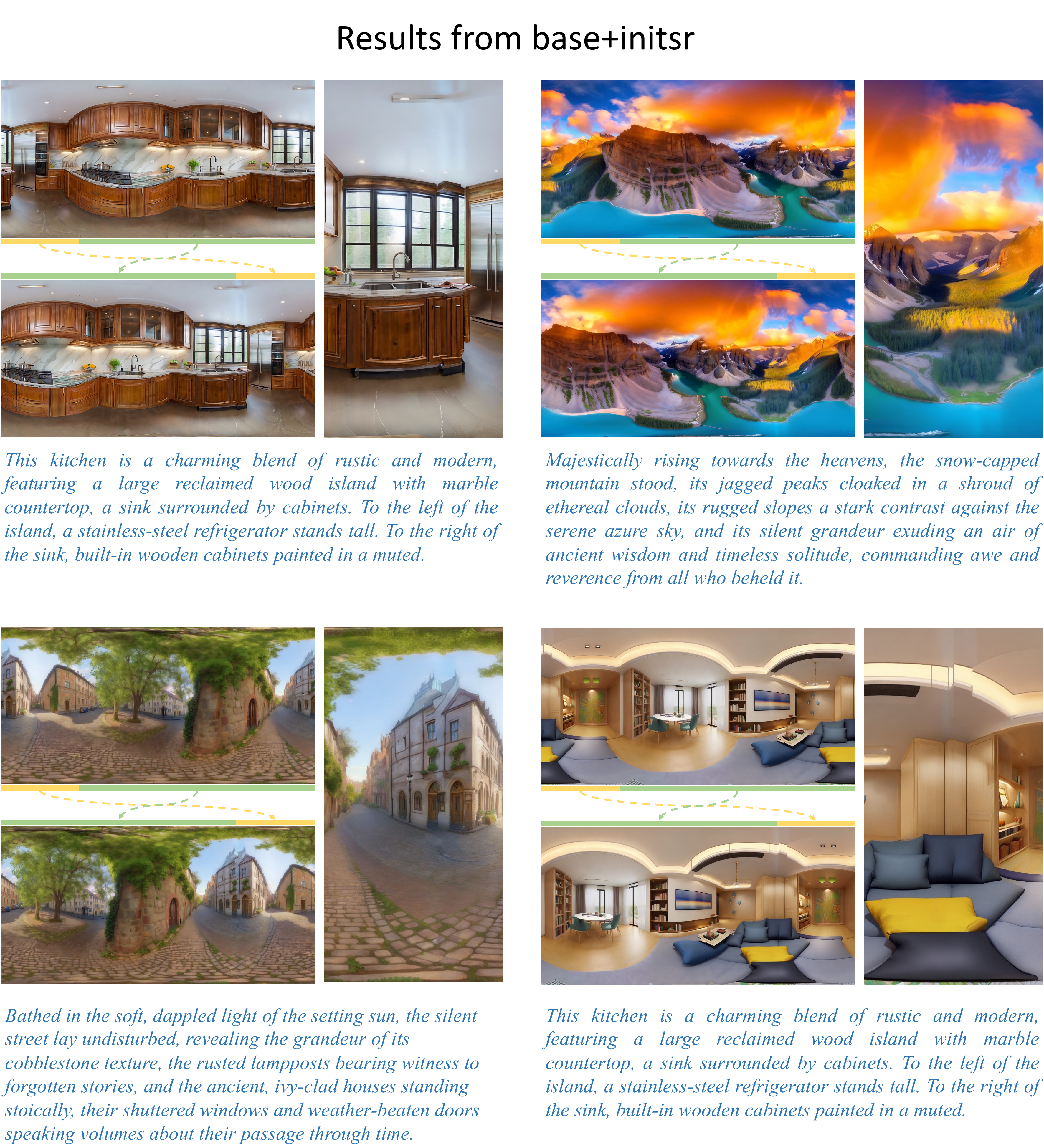}
  \caption{Results from Base+InitSR.}
  \label{samples-base+initsr}
\end{figure*}

\begin{figure*}
  \centering
  \includegraphics[width=\linewidth]{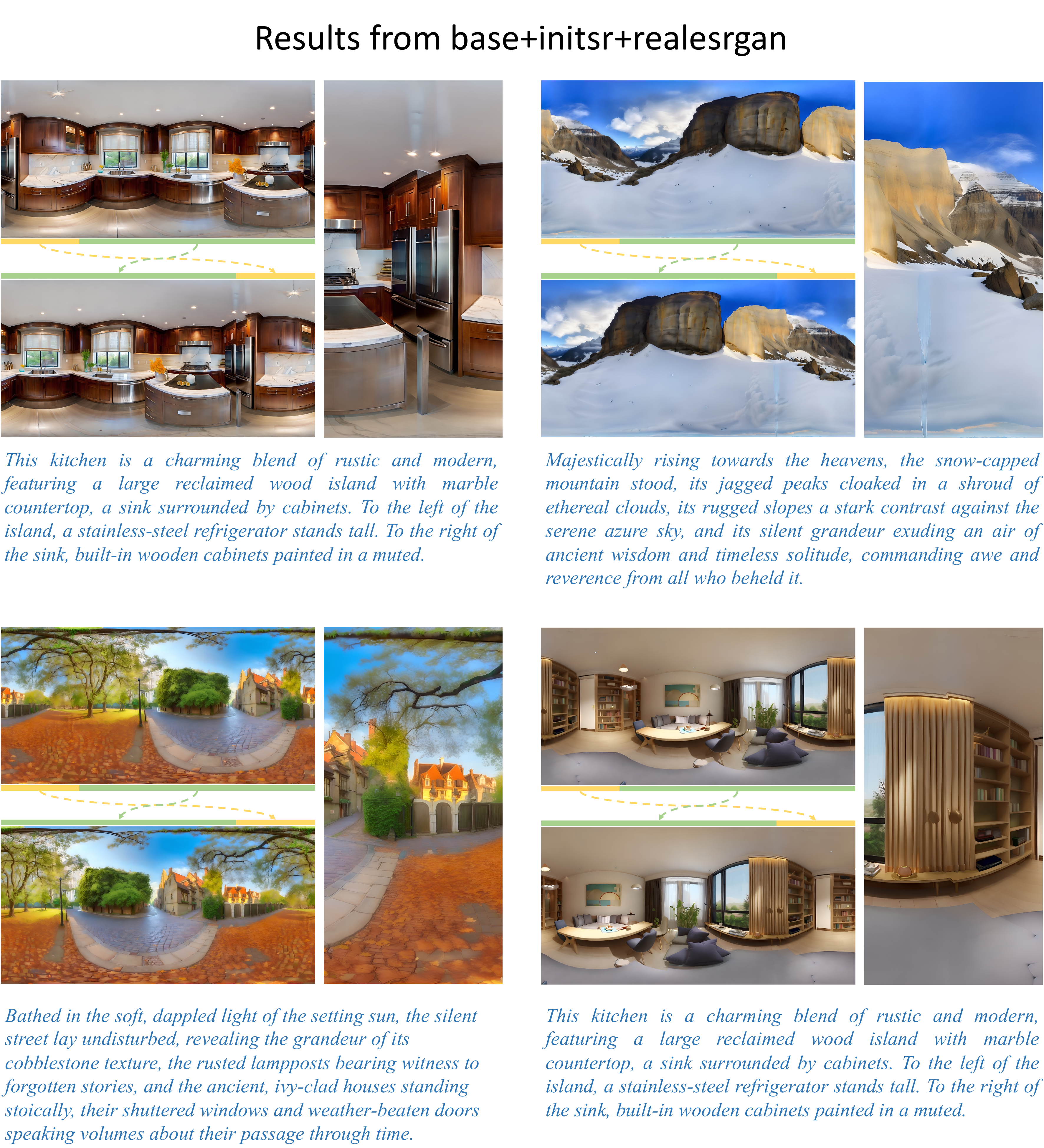}
  \caption{Results from Base+InitSR+ReslESRGAN. It can be observed that, the geometric continuity of the rightmost and the leftmost sides of our results are smooth and nearly no cracks. Some artifacts in the top two rows are cost by the RealESRGAN.}
  \label{samples-base+initsr+reslesrgan}
\end{figure*}

\begin{figure*}
  \centering
  \includegraphics[width=\linewidth]{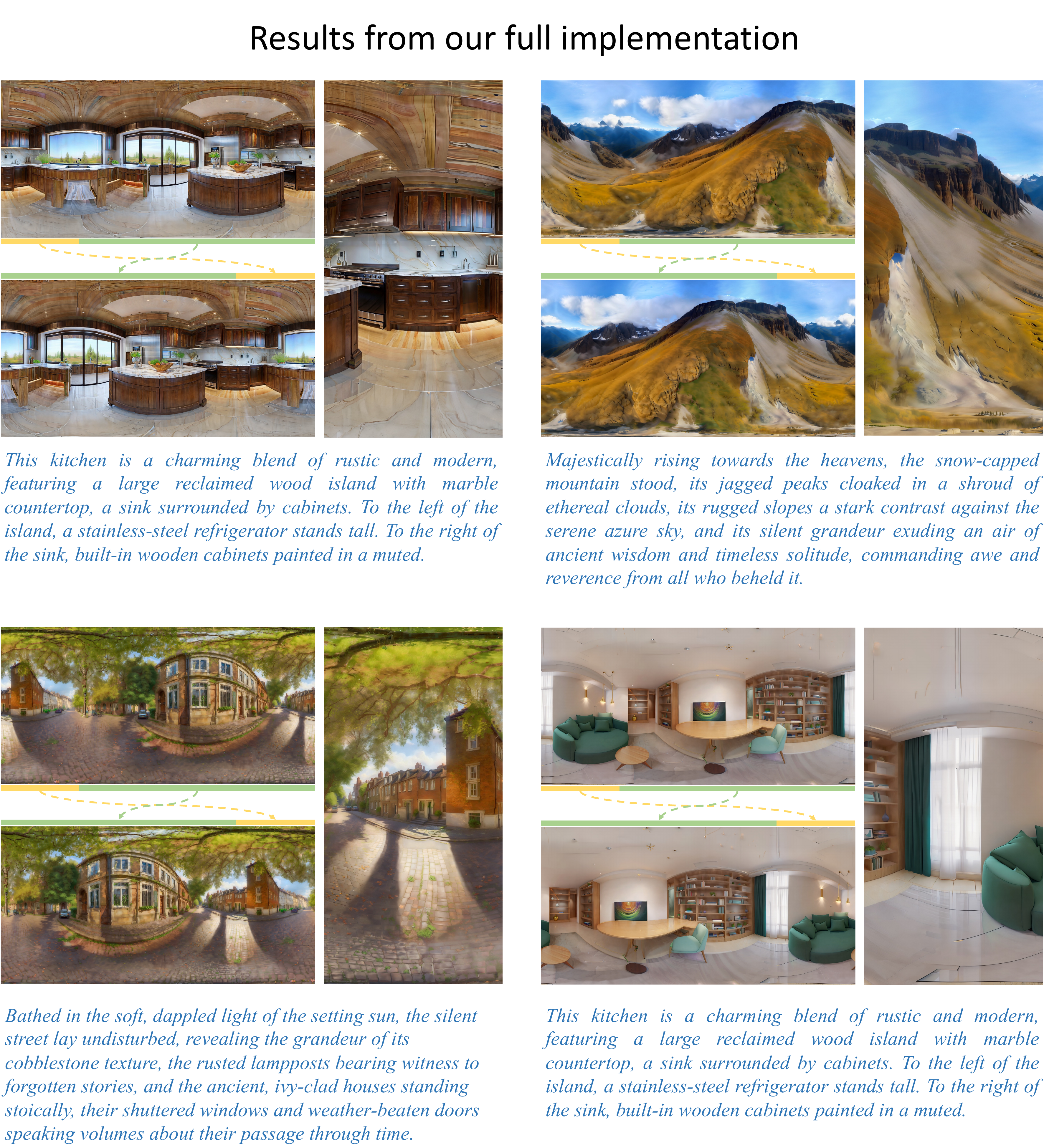}
  \caption{Results from the full implementation.}
  \label{samples-fullimp}
\end{figure*}

We show some testing results at different stages of the Text-to-360-Panoramas task in Fig.~\ref{samples-base}, Fig.~\ref{samples-base+initsr}, Fig.~\ref{samples-base+initsr+reslesrgan}, and Fig.~\ref{samples-fullimp}. The input prompts it fetch at the MVDiffusion project page (\url{https://mvdiffusion.github.io/})


\section{Limitations}
The base model is trained using the DreamBooth~\cite{ruiz2022dreambooth} technique, so it can not be changed with the models from CIVITAI (\url{https://civitai.com/}) for stylizing purposes. Adding some style descriptions (such as 'cartoon style' and 'oil painting style') in the prompt does not work. One can generate an initial 360 image using our method, and then use ControlNets (like canny and depth) with different base models to change the style.

\clearpage

{
    \small
    \bibliographystyle{ieeenat_fullname}
    \bibliography{main}
}


\end{document}